\documentclass[letterpaper, 10 pt, conference]{ieeeconf}  % Comment this line out if you need a4paper

\IEEEoverridecommandlockouts                              % This command is only needed if 
\usepackage{graphics} % for pdf, bitmapped graphics files
\usepackage{epsfig} % for postscript graphics files
\usepackage{mathptmx} % assumes new font selection scheme installed
\usepackage{times} % assumes new font selection scheme installed
\usepackage{amsmath} % assumes amsmath package installed
\usepackage{amssymb}  % assumes amsmath package installed
\usepackage{multirow}
\usepackage{adjustbox}
\usepackage{booktabs}
\usepackage{caption}
\usepackage{cite}
\usepackage{array}
\usepackage{siunitx}
\usepackage{xcolor}
\usepackage{color,soul}
\usepackage{float}
\usepackage{stfloats}
\usepackage{graphicx}
\usepackage{hyperref}
\usepackage{subcaption}

\title{\LARGE \bf
Precision Harvesting in Cluttered Environments: Integrating End Effector Design with Dual Camera Perception
}

\author{Kendall Koe$^{1}$, Poojan Kalpeshbhai Shah$^{2}$, Benjamin Walt$^{2}$, Jordan Westphal$^{2}$, Samhita Marri$^{3}$, \\Shivani Kamtikar$^{1}$, James Seungbum Nam$^{2}$, Naveen Kumar Uppalapati$^{4}$, Girish Krishnan$^{2}$, Girish Chowdhary$^{1}$ \thanks{$^{1}$Computer Science, $^{2}$Mechanical Science and Engineering, $^{3}$Electrical and Computer Engineering, $^{4}$National Center for Supercomputing Applications, UIUC, USA. {\tt(kfkoe2, pks11, walt, jwest33, marri2, skk7, sn29, uppalap2, gkrishna, girishc)@illinois.edu} \newline \indent This work is supported by funding from the USDA-NIFA UIE grant (2023-70019-39365) and the AIFARMS National AI institute in agriculture, backed by AFRI grant no. 2020-67021-32799/project accession no.1024178 from USDA NIFA. Additionally, it benefits from a joint NSF-USDA COALESCE grant 2021-67021-34418. Furthermore, support is provided by the NSF grant No. DBI-2019674 for CROPPS.}}

\begin{document}

\maketitle
\thispagestyle{empty}
\pagestyle{empty}
%%%%%%%%%%%%%%%%%%%%%%%%%%%%%%%%%%%%%%%%%%%%%%%%%%%%%%%
%%%%%%%%%%%%%%%%%%%%%%%%%
\begin{abstract}
    Due to labor shortages in specialty crop industries, a need for robotic automation to increase agricultural efficiency and productivity has arisen. Previous manipulation systems perform well in harvesting in uncluttered and structured environments. High tunnel environments are more compact and cluttered in nature, requiring a rethinking of the large form factor systems and grippers. We propose a novel codesigned framework incorporating a global detection camera and a local eye-in-hand camera that demonstrates precise localization of small fruits via closed-loop visual feedback and reliable error handling. Field experiments in high tunnels show our system can reach an average of 85.0\% of cherry tomato fruit in 10.98s on average. 

\end{abstract}
%%%%%%%%%%%%%%%%%%%%%%%%%%%%%%%%%%%%%%%%%%%%%%%%%%%%%%%%%%%%%%%%%%%%%%%%%%%%%%%%
\section{INTRODUCTION}
Decreasing food miles and increasing sustainable agricultural practices have prompted interest in urban agriculture in recent years. High Tunnels (HTs) \cite{Janke2017,Lamont2009} are low-cost, unheated, metal-tube structures covered with greenhouse plastic to create a protected environment for crops, and are ideal for urban farming. However, automated crop harvesting within urban high tunnels poses a challenge due to the tight spacing between rows of plants \cite{demchak2009small,kootstra2021selective,bruce2019farmers}. For example, the USDA recommends that rows of tomatoes planted in high tunnels be spaced between 91cm and 122cm apart, which leaves even less space once plants mature \cite{nrcs2014planting}. This limits large form-factor systems from operating in these constrained and dense environments. Limited form factor further restricts power and computation capabilities, requiring a simple yet robust framework for perception functionalities such as continuous detection of target berries and planning for manipulation. This paper highlights how nuances in the hardware design and vision-based perception can impact the success of harvesting in a constrained, cluttered high-tunnel environment.

\begin{figure}[htbp]
    \includegraphics[width=\linewidth]{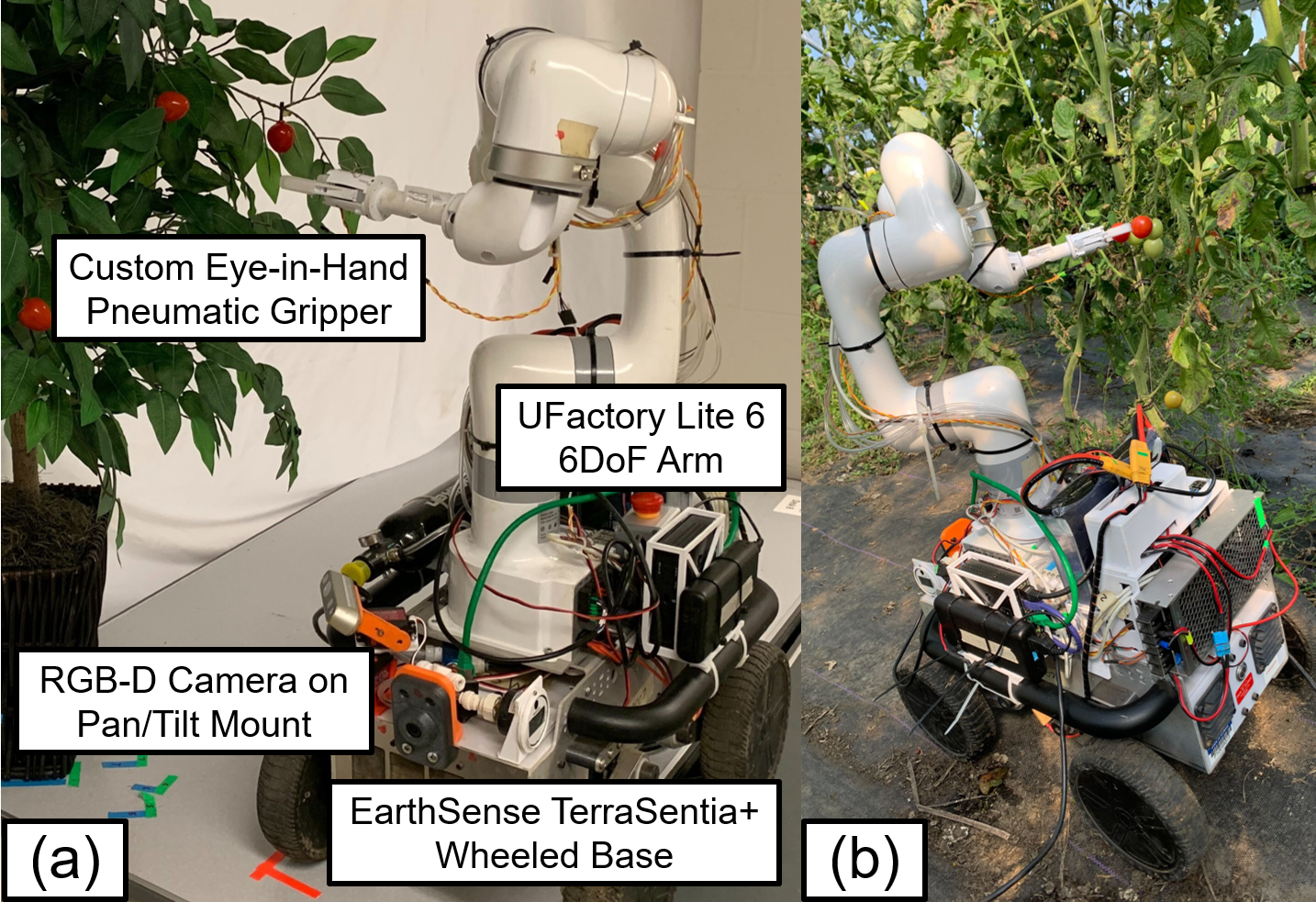}
    \caption{Robot picking cherry tomatoes with our Detect2Grasp framework, (a) from an artificial plant and (b) in an outdoor high tunnel. The berry is located by the RGB-D base camera, and then an admissible approaching pose is computed for the manipulator. The manipulator moves to this pose and then uses the gripper camera to visually servo to grasp the fruit.}
    \label{fig:system_diagram}
    \vspace{-20pt}
\end{figure}

\subsection{Related work}
Current state-of-the-art robotic platforms that harvest fruits in orchards and greenhouses typically consist of a mobile platform and a manipulator with a gripper end effector \cite{xiong2020autonomous, li2023hybrid, li2022hybrid, bac2017performance, lili2017development,  feng2018design, zhang2022algorithm, lehnert2018sweet}. Fruits or berries are detected using multiple RGB or depth cameras, some placed on the mobile platforms and others placed at the distal manipulator end. A combination of path planning and visual servoing is used to approach the target fruit and close in on the fruit respectively.  

Robust harvesting requires codesign of the perception pipeline and the manipulation hardware for a given setting. If berries are visible and are on the periphery of the vine, which usually occurs in commercial settings where defoliating occurs prior to harvest \cite{gao2022development}, a single camera placed at the distal end of the manipulator can be sufficient to detect, segment, and servo the end effector to the berry. \cite{lili2017development} and \cite{feng2018design} each develop a robot for harvesting tomatoes. However, both systems are larger than 122cm in width and use binocular stereo cameras at the end effector.
In unstructured and cluttered under-canopy settings, the perception pipeline is more involved. Reconstruction of the fruit and a visual map of the obstacles aid in planning a non-intrusive path with minimal to no collisions towards the fruit \cite{magistri2024efficient, gibbs2019active}. Herein lies a trade-off between the computational requirement of the system and the hardware. Most existing systems employ a distal depth camera at close proximity to the end effector or gripper. If the placement of the camera is too close to the end effector, it may have a restricted field of view and may collide with obstacles, thus limiting reach. On the other hand, if the camera is placed a distance away, its field of view may not include the target during the reach, which limits the effectiveness of visual servoing due to target localization errors \cite{feng2018design, lili2017development, jiang2022development}. Some existing systems increase the gripper size to handle these errors \cite{li2024recognition}. Others use multiple robot arms, one for perception and the other for manipulation \cite{lenz2024hortibot, stavridis2022bimanual}, but this increases the cost of the system. There is thus a gap for a platform that maximally combines intentional hardware improvements and computationally less intense perception for robust berry harvesting in the high tunnel settings.

\subsection{Approach}

In this paper, we designed a harvesting robot with a compact mobile base ($\sim$35cm width) and a 6-DOF manipulator arm with a slender distal end. The system consists of two cameras, an RGB-D camera stationed on the mobile base for a global scan, and a small-form factor RGB camera at the end effector. An important feature of the design is to collocate the small-form factor RGB camera between the fingers of a pneumatic gripper. Furthermore, we present Detect2Grasp, a strategy to grasp small to medium-sized fruit (such as cherry tomatoes) by leveraging the combination of the two cameras. We show that localizing and grasping can be done solely with a small RGB camera once the target is in view, eliminating the need for high-fidelity, yet bulky sensors at the distal end. Although we trade off depth and high-resolution image information for compactness, we demonstrate reliable grasping capability across a variety of test environments. The end effector navigates to an initial pose estimated by the base depth camera and performs a local search around this pose, if needed, to ensure the target fruit is within the field of view. Then, a visual servoing algorithm iteratively centers the fruit and proceeds forward until the berry is within the grippers. Extensive evaluations in both lab and field settings are conducted to characterize the system's accuracy and robustness. Specifically, in a cluttered under-canopy setting, we demonstrate that a collocated RGB camera at the end effector is more effective for reaching target fruit than placing high-resolution cameras away from the gripper. 
\section{HARDWARE DESIGN}

\begin{figure}[htp]
    \centering
    \captionsetup[subfigure]{aboveskip=2pt}
    \begin{subfigure}[t]{0.4\linewidth}
        \includegraphics[width=\linewidth]{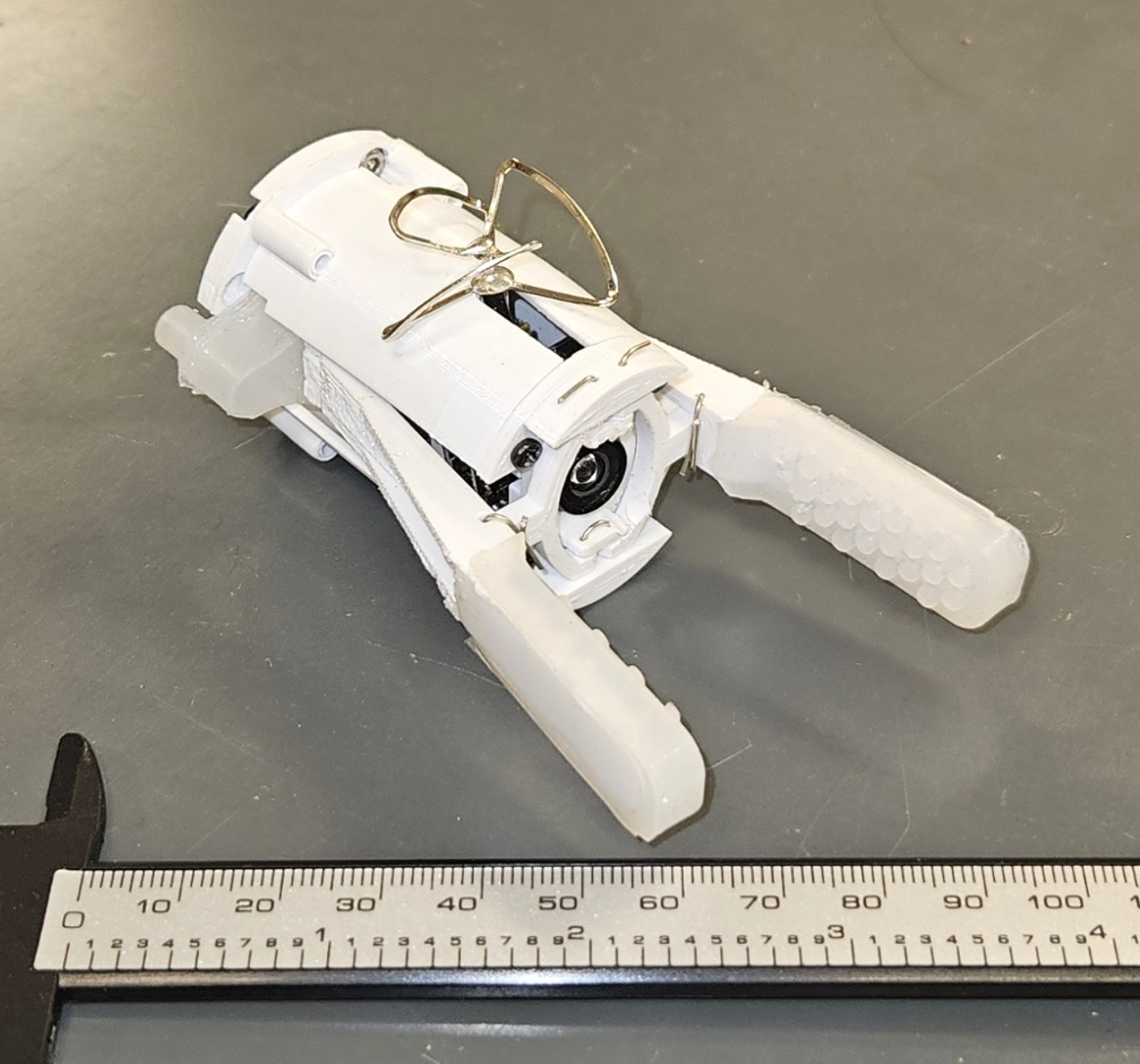}
        \caption{Physical gripper}
        \label{fig:real_gripper}
    \end{subfigure}
    \hspace{5pt}
    \begin{subfigure}[t]{0.4\linewidth}
        \includegraphics[width=\linewidth]{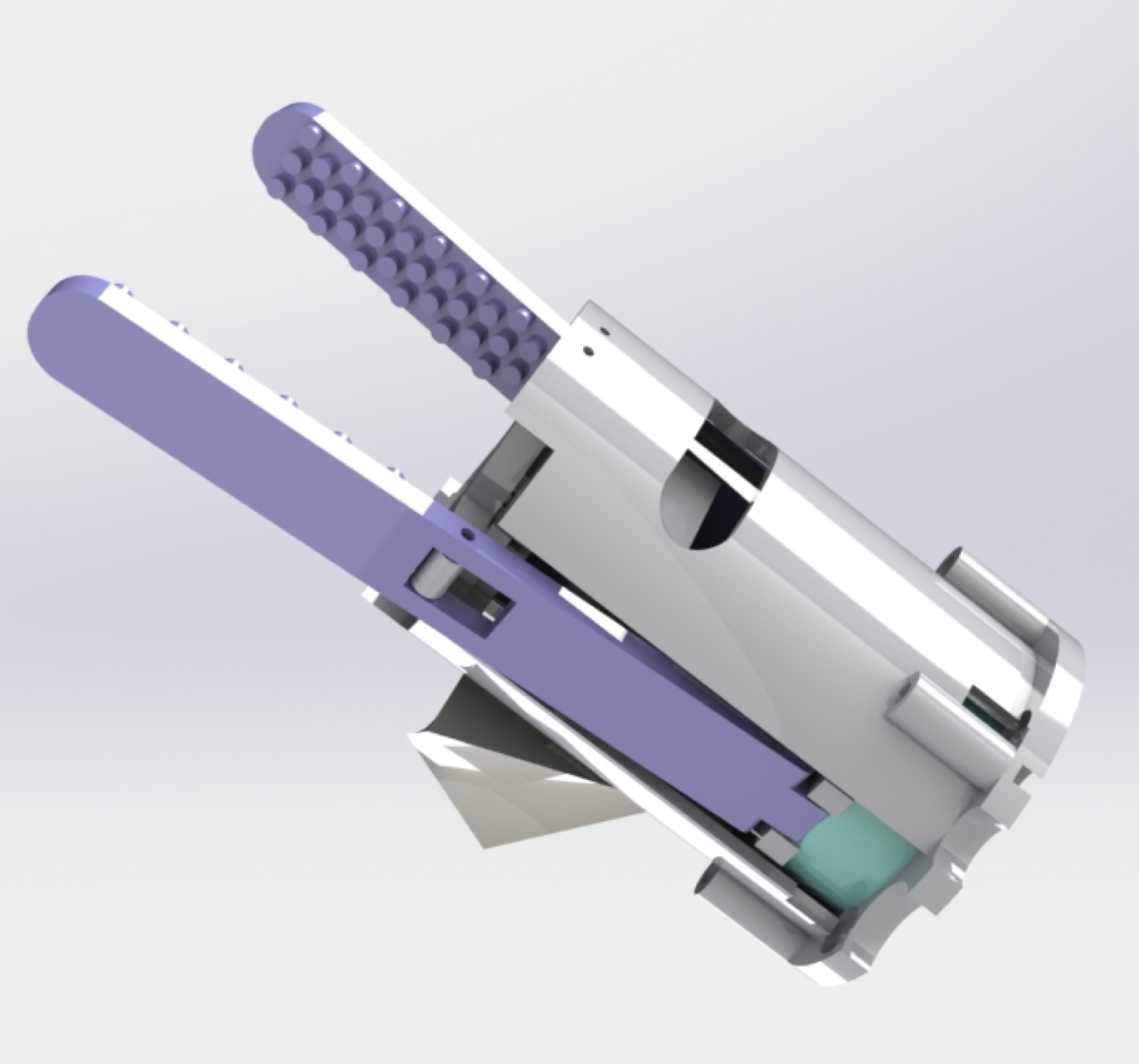}
        \caption{Assembled view}
        \label{fig:gripper_assembled}
    \end{subfigure}
    \begin{subfigure}[t]{0.85\linewidth}
        \includegraphics[width=\linewidth]{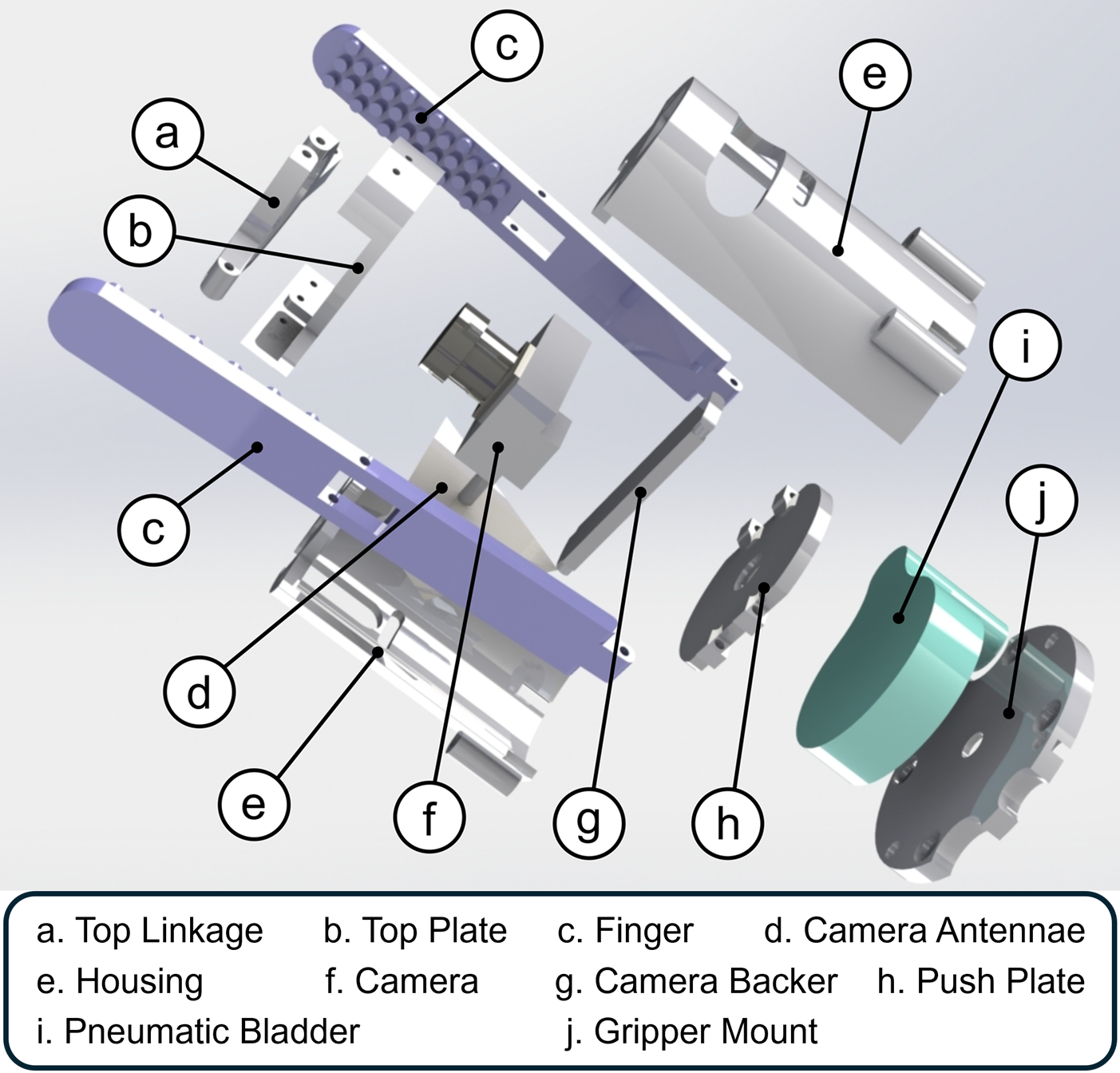}
        \caption{Exploded and annotated view of model}
        \label{fig:gripper_exploded}
    \end{subfigure}
    \caption{Our custom pneumatic gripper with a camera collocated with the central axis.}
    \label{fig:Gripper}
    \vspace{-20pt}
\end{figure}

Previous works describe localizing fruit between grippers as a main failure mode \cite{yaguchi2016development, gao2022development}. Without visual or tactile feedback, systems rely on open-loop controls to achieve a precise grasp assuming a fixed fruit location. Cameras without direct sight of the grippers have a blind spot, while those with a direct line of sight have the gripper obstructing the field of view. Additionally, cameras placed at offset have a greater chance of colliding with foliage due to the larger profile, especially in cluttered environments. Three-fingered grippers or fin ray grippers also have a larger profile but offer more security when grasping \cite{Crooks2016}.

Our design is tied to its use case in dense, cluttered high tunnel environments. We collocate a camera between pneumatic grippers hypothesizing that this design, in combination with our visual servoing algorithm, would enable precise closed-loop manipulation and reduce collisions with the environment by maintaining visual contact with the grippers continuously while having a small profile. The gripper is a constrained four-bar linkage that closes when the pneumatic bladder is inflated. The inner surfaces of the fingers are lined with Dragon Skin 20, a compliant silicone rubber compound, to minimize damage to the fruit. The gripper's stroke was designed for harvesting small fruit requiring a small form factor camera (AKK BA2) for collocation. Details are seen in Figure \ref{fig:Gripper}.

Additionally, to avoid singularities resulting from the alignment of two revolute joints in our robotic arm, a 90-degree bend was designed in the distal link. In order to balance workspace size, minimize potential environment interactions, and end effector stability, a link length of 8 cm was chosen. The remainder of the experimental hardware consists of a UFactory Lite 6 arm mounted on an EarthSense TerraSentia+ mobile base as seen in Figure \ref{fig:system_diagram}. The mobile base has been modified to provide power and compressed air to operate the arm and gripper in the field. An Intel Realsense D435i RGB-D camera is mounted on a servo-controlled pan and tilt system located on the mobile base. 

\section{SOFTWARE DESIGN}

\begin{figure*}[htp!]
\centering
\smallskip
  \includegraphics[width=0.95\textwidth]{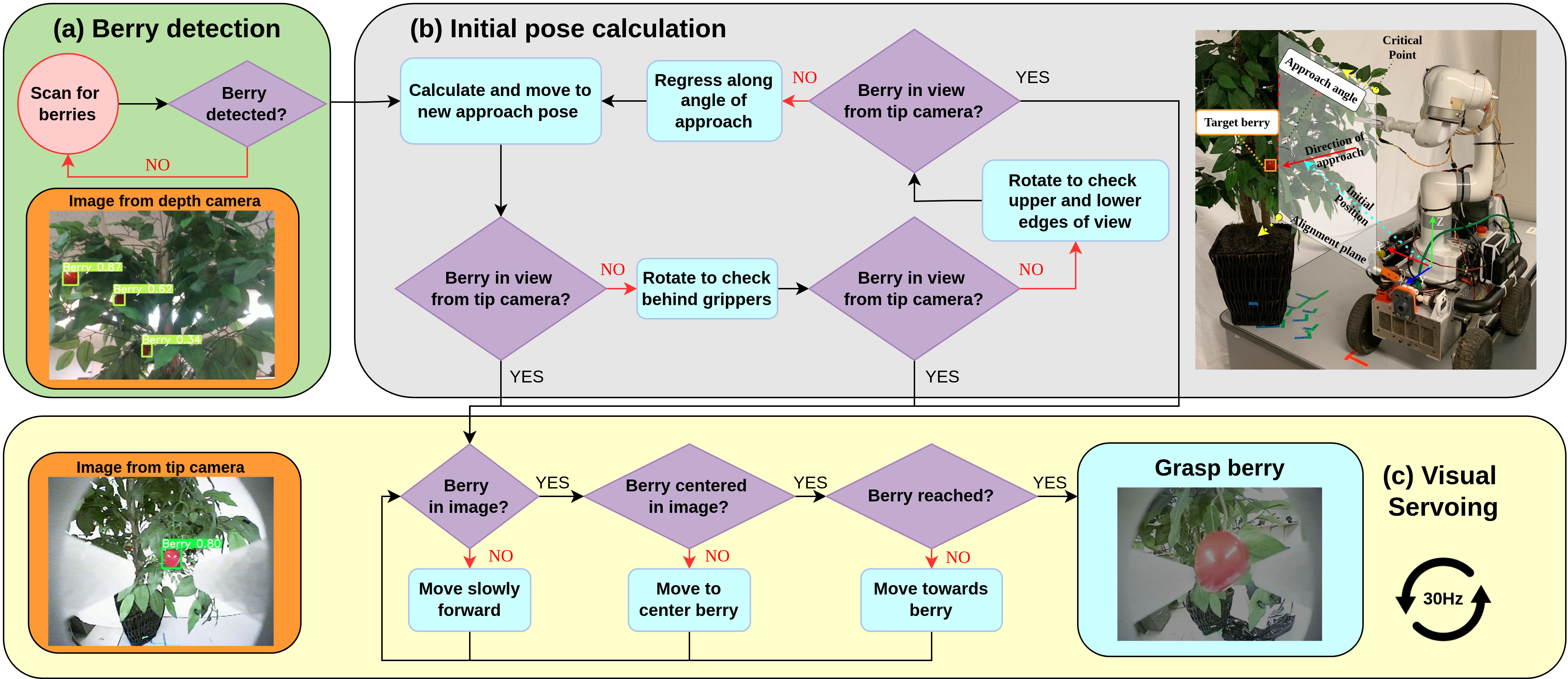}
  \caption{Flow diagram of our Detect2Grasp algorithm that includes berry detection, initial pose calculation, and visual servoing phases: (a) Using the RGB-D camera to scan the area and identifying berries with a Yolov7 detector. (b) Establishing a plane from the depth camera to the detected berry, aligning an initial pose as detailed in Section \ref{sec:compute_initial_pose}, and then moving the manipulator to this pose, recomputing if the berry isn't visible. (c) With the berry visible, it proceeds to visual servoing, employing a Cartesian velocity controller for approach, maintaining berry centering with an inner loop, and defining 'reached' when the berry significantly fills the image.}

  \label{fig:Visual_servoing_flow}
  \vspace{-.3cm}
\end{figure*}

Our Detect2Grasp algorithm consists of 3 parts: detecting berry targets, calculating the initial pose, and visual servoing as seen in Figure \ref{fig:Visual_servoing_flow}. Detect2Grasp is built using a state machine developed with the Smach-ROS Python library. A state machine highlights clarity and modularity by simplifying the control of repeated, sequential, and conditional tasks into predetermined function transitions. This also helps enable collaborative software development.   

\subsection{Fruit Detection}
\label{sec:fruit_detection}
To reach and grasp the targets of interest, detection becomes key in achieving this goal. Detection was performed using Yolov7 \cite{wang2023yolov7}, a neural network that enables real-time bounding box detection on RGB images on a GPU. A total of 126 images were collected from our tip camera in a lab setting and labeled for fine-tuning the fruit detection model as shown in Figure \ref{fig:system_diagram}. To increase detection quality, image flips, color augmentations and other training hyperparameters were configured following the provided recommendations \cite{wang2023yolov7}. This detection module was used on the base and tip camera. 

To estimate the initial 3D location of the target fruit, we overlay the RGB image with the depth image to obtain the depth readings corresponding to the fruit detected by the Yolov7 model. Although this model is trained solely with indoor images of artificial fruit, we observed that it is able to detect real cherry tomatoes. To aid in transferring this model to outdoor environments, experiments were conducted to determine optimal light exposure settings for the RGB-D camera under these conditions.

\subsection{Computing Initial Pose}
\label{sec:compute_initial_pose}
To compute the initial pose for the manipulator, the plane that contains the vector connecting the depth camera center to the berry and the vertical axis is found. A portion of this plane is assumed to be obstacle-free, as the depth camera has already detected the berry within it. 

A prior calibration of the approach angle is done by performing a workspace analysis. First, we find desirable approach angles for selected points on the boundary of the manipulator's workspace. Then, for the target point, which would be contained by the convex hull of the boundary points, we linearly interpolate between pairs of boundary approach angles to find the desired approach angle. 

A position along this line of approach, offset from the berry position, is used as the initial position to increase the probability of seeing the berry. The ROS MoveIt package is used to achieve the initial pose. If the berry is not detected, then rotations left and right are also checked to ensure the berry is not behind the gripper. Additionally, rotations upward and downward within the plane are given to search for the berry. To ensure robustness should the berry still not be visible, the offset would increase and the process repeated. Additional candidate poses are tried until the target can be detected from the tip. 
This is visualized in Figure \ref{fig:Visual_servoing_flow}b.

\subsection{Visual Servoing}
\label{sec:Visual_Servoing_Algorithm}
Once an initial pose with the berry in view is reached, the visual servoing algorithm is executed. First, the berry center is moved to the image center with a small dead band around the center to prevent oscillations. 

To center the berry, the direction vector connecting the center of the image to the center of the berry’s bounding box is found. This direction vector is the reference trajectory that is tracked by a Proportional-Integral-Derivative controller with an anti-windup and frequency of 30Hz, limited by the camera streaming rate.

After centering, the system stops and then proceeds towards the berry along the line of sight of the end effector. Should the berry center ever leave the predefined image center, the centering control loop would again be executed. If the berry is occluded mid-algorithm, the manipulator moves forward slowly in an attempt to find the berry. 

A heuristic determines the stopping criteria for the algorithm based on the berry's size in image space. The berry must be within the grippers to exceed the threshold for determining a successful reach. The collocated camera enables this stopping criterion.

This algorithm assumes that berry sizes are similar, but the same algorithm can be used for different-sized fruits by re-calibrating thresholds and swapping out the gripper accordingly. During the visual servoing, the system may detect multiple berries simultaneously. In this scenario, the berry closest to the center of the image is prioritized, enabling convergence to the target berry with stable detections. 

\begin{table*}[htp!]
  \begin{center}
  \smallskip
  \renewcommand{\arraystretch}{1.2}
  \resizebox{\textwidth}{!}{
  \begin{tabular}{c c c c}
    \toprule
    \multirow{2}{*}{{\textbf{Test Scenario}}} & \multirow{2}{*}{\textbf{Success Rate (\%)}} & \multirow{2}{*}{\textbf{Average Time to Reach Berry (s)}} 
     & \multirow{2}{*}{\textbf{Number of Experiments}} 
    \\ \\
    \midrule
    
     Base Depth Camera Only & 0.0 & -- & 15
     \\
    Base VS on Artificial Plant & 87.5 & 9.35 $\pm$ 3.58 & 40
    \\ 

    + Corrupted Depth & 46.7 & 8.55 $\pm$ 1.12 & 15 
    \\
    
    + 13x Light Intensity & 60.0 & 8.84 $\pm$ 2.94 & 15
     \\ 

    + 20x Light Intensity & 80.0 & 11.89 $\pm$ 7.55 & 15
     \\
     Hanging Vine Environment & 86.7 & 9.28 $\pm$ 1.99 & 15
     
     \\
     Distal Depth Camera & 60 & 11.66 $\pm$ 4.10 & 15  
     \\
     Outdoor High Tunnel & 85.0 & 10.98 $\pm$ 5.81 & 20
     \\
    \bottomrule
    \end{tabular}}
  \caption{This table presents a summary of the experiment results of our method used in the lab and field, along with results under adversely altered environments and additional sensor noise. Results from ablation studies that remove components are also included.}
  \label{baseline_ablation_tests}
  
  \end{center}
  \vspace{-.5cm}
\end{table*}  

\begin{figure}[htp]
\centering
\includegraphics[width=0.95\linewidth]{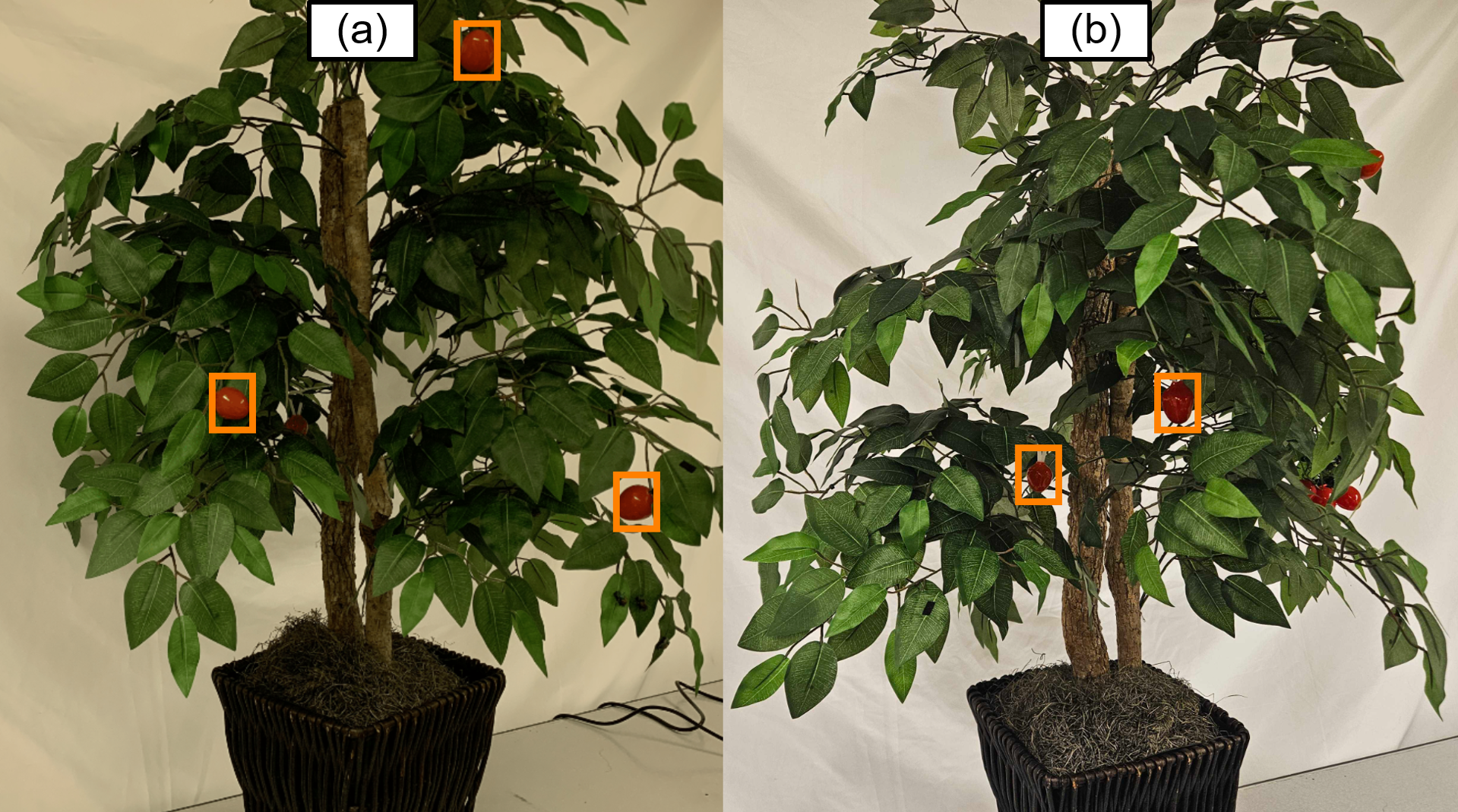}
  \caption{The berries on (a) the periphery and (b) under the canopy are used for testing. Peripheral berries are exposed outward while berries under the canopy are recessed towards the stem of the plant. Berries detected by the base camera are marked with orange.}
  \label{fig:Berry_Positions}
  \vspace{-10pt}
\end{figure}
\section{EXPERIMENTS}
%In this section we conduct detailed experimental tests to evaluate the effectiveness of Detect2Grasp framework in several realistic scenarios.
\begin{figure*}[htp!]
\centering
\smallskip
\includegraphics[width=0.98\textwidth]{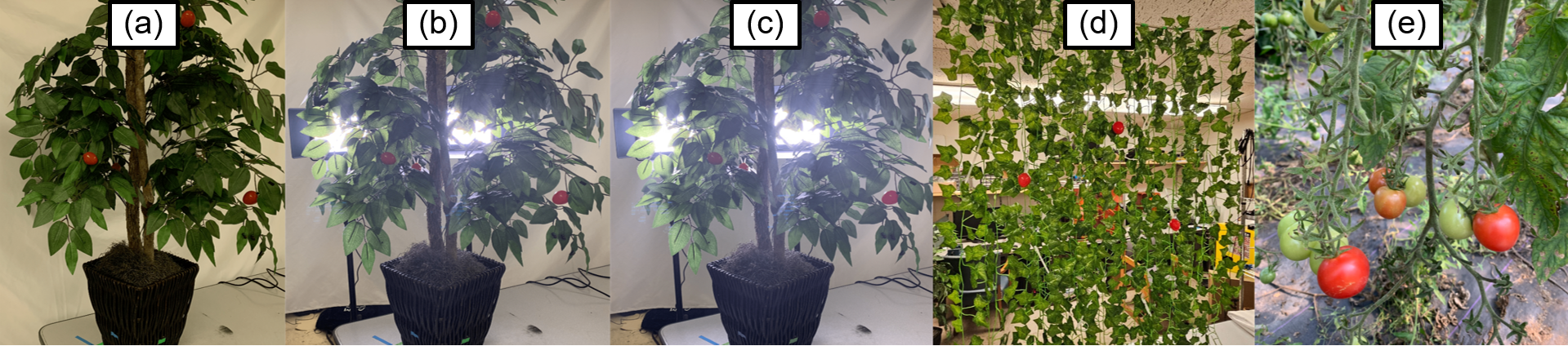}
  \caption{Experimental Setups. (a) Base Setup (b) 13x Lighting (c) 20x Lighting (d) Hanging Vine (e) Outdoor High Tunnel}
  \label{fig:experimental_setups}
  \vspace{-15pt}
\end{figure*}
\subsection{Base Setup}

We conducted experimental tests to evaluate the effectiveness of our system in several realistic scenarios. To validate the need for a camera at the distal end of the system, we initially rely solely on RGB-D camera measurements. Precise manipulation from distance is shown to be infeasible in cluttered environments. Errors accumulate from stereo matching, poor lighting conditions, sensor degradation, and other sources resulting in an average error of 6.8cm between the berry and center of the gripper over 15 trials. This validates the need for an additional camera integrated into the distal end of the manipulator.

To establish a baseline, 40 berry positions were tested with our collocated gripper. To simulate outdoor high tunnel dimensions, the berries were placed between 30cm and 45cm at varying heights surrounded by foliage. 60\% of the berries were on the periphery while 40\%  were under the canopy as seen in Figure \ref{fig:Berry_Positions}. The testing environment can be seen in Figure \ref{fig:experimental_setups}a and the results from the test are summarized in Table \ref{baseline_ablation_tests}. Our system successfully reaches the target with 87.5\% success in 9.35 seconds on average with failures arising from occlusions and workspace limitations when servoing (see Figure \ref{fig:Failure_Cases}).

\subsection{Corrupted Depth Tests}
To evaluate the robustness of Detect2Grasp, zero-mean Gaussian noise was added to the given target position with a standard deviation of 7.5cm. This simulates sensor degradation and further inaccuracies from the depth camera. Our system achieved a 73.3\% success rate over 15 trials with an average reaching time of 8.55s, even overcoming a noise reading corrupted by almost 20cm. This lower reaching time and variance is due to the highly noisy poses resulting in failure. Primarily, the corrupted initial pose induced poor approach angles which resulted in a failure due to the limited workspace.

\subsection{Lighting Tests}
To evaluate the robustness of our berry detector, studio lights were used to increase the light intensity. With a baseline reading of 190lux at the depth camera, additional lighting scenarios with 2470lux ($\sim$13x) and 3780lux ($\sim$20x) were evaluated as seen in Figure \ref{fig:experimental_setups}b and \ref{fig:experimental_setups}c. 15 berries at each lighting intensity were tested. The average reaching time with 13x light intensity was 8.84s with 80\% success. With 20x light intensity, the average reaching time was 11.89s with 80.0\% success. Performance is slightly degraded because of increased detection failures (seen as orange in Figure \ref{fig:Failure_Cases}), but shows robustness to illumination changes.

\subsection{Hanging Vine Environment}

To evaluate the model on a different setup, an artificial hanging vine setup was created as seen in Figure \ref{fig:experimental_setups}d. In 15 attempts of reaching berries in this environment, 13 were successful with an average reaching time of 9.28s with a standard deviation of 1.99s. Even with different surrounding environments, the target fruit was still reached.

\subsection{High Tunnel Experiments}
Our system was also evaluated in a local high tunnel. 20 tests conducted mid-morning, our system showed reliable transfer between environments, robustness to outdoor lighting, and successful reaching of real fruit. An 85\% success rate was achieved with an average reaching time of 10.98 seconds.

\begin{figure}[htp!]
    \centering
    \captionsetup{justification=centering}
    \captionsetup[subfigure]{aboveskip=2pt}
    \begin{subfigure}[t]{0.47\linewidth}
        \includegraphics[width=\linewidth]{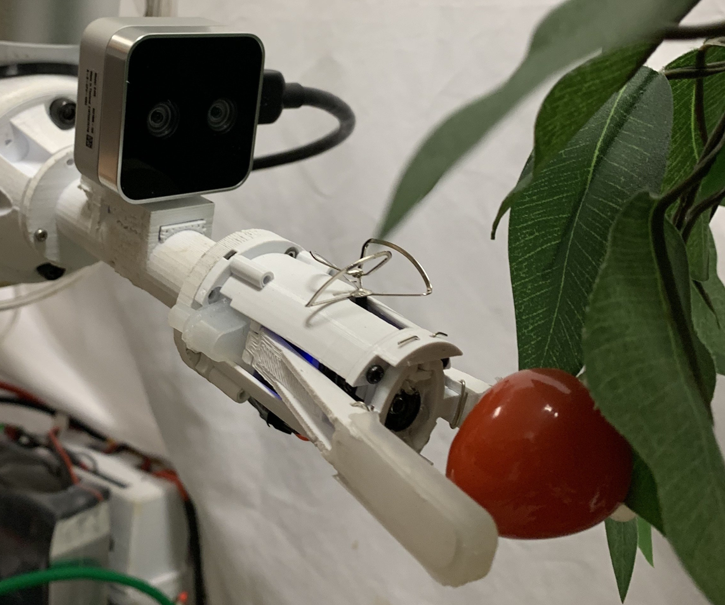}
        \caption{Successful grasp of peripheral berry}
        \label{fig:distal_config}
    \end{subfigure}
    \hspace{2pt}
    \begin{subfigure}[t]{0.47\linewidth}
        \includegraphics[width=\linewidth]{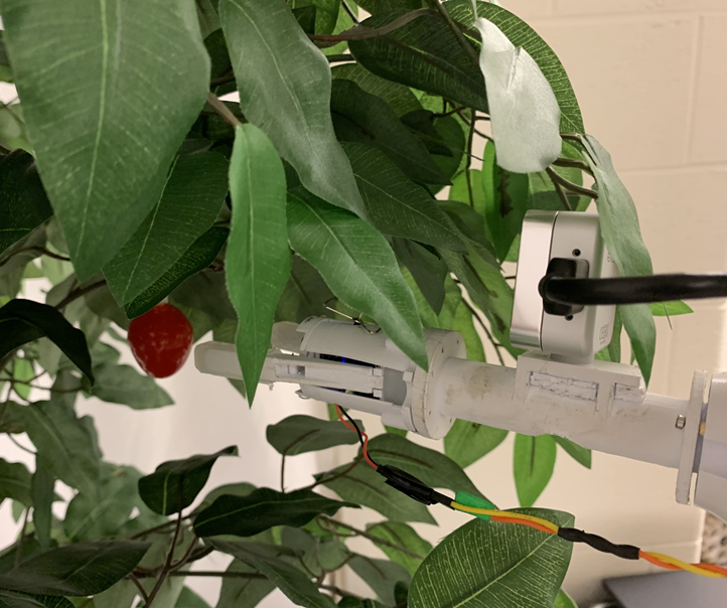}
        \caption{Failure to reach under canopy berry}
        \label{fig:distal_failure}
    \end{subfigure}

  \captionsetup{justification=raggedright}
  \caption{The distal depth camera setup used as a baseline. In experiments with this setup, the RGB camera enclosed in the gripper was not used. This gripper successfully grasps berries on the periphery (a), but more often collides with the canopy due to the larger profile (b).}
  \label{fig:distal_depth_setup}
  \vspace{-10pt}
\end{figure}

\subsection{Distal Depth Camera}

To compare more directly to systems with depth cameras offset from the gripper like \cite{kondo2010development, yaguchi2016development, lili2017development, feng2018design, gao2022development}, experiments were done with a distal depth camera, specifically the Intel RealSense D405, attached on the final link of the manipulator as seen in Figure \ref{fig:distal_depth_setup}. Instead of using the RGB camera to center the berry, once the initial pose was achieved, a trajectory passing through the position 4cm offset from the berry (the length of the grippers) was passed through prior to reaching the berry between the grippers. This lessened the chances of colliding directly with the berry or other parts of the plant while approaching the fruit. This method achieved 77.8\% success rate on peripheral berries but decreased to 60\% overall when including under-canopy berries. This is due to increased chances of collisions due to the larger profile under the canopy.

\begin{figure}[htp]
\centering
  \includegraphics[width=\columnwidth]{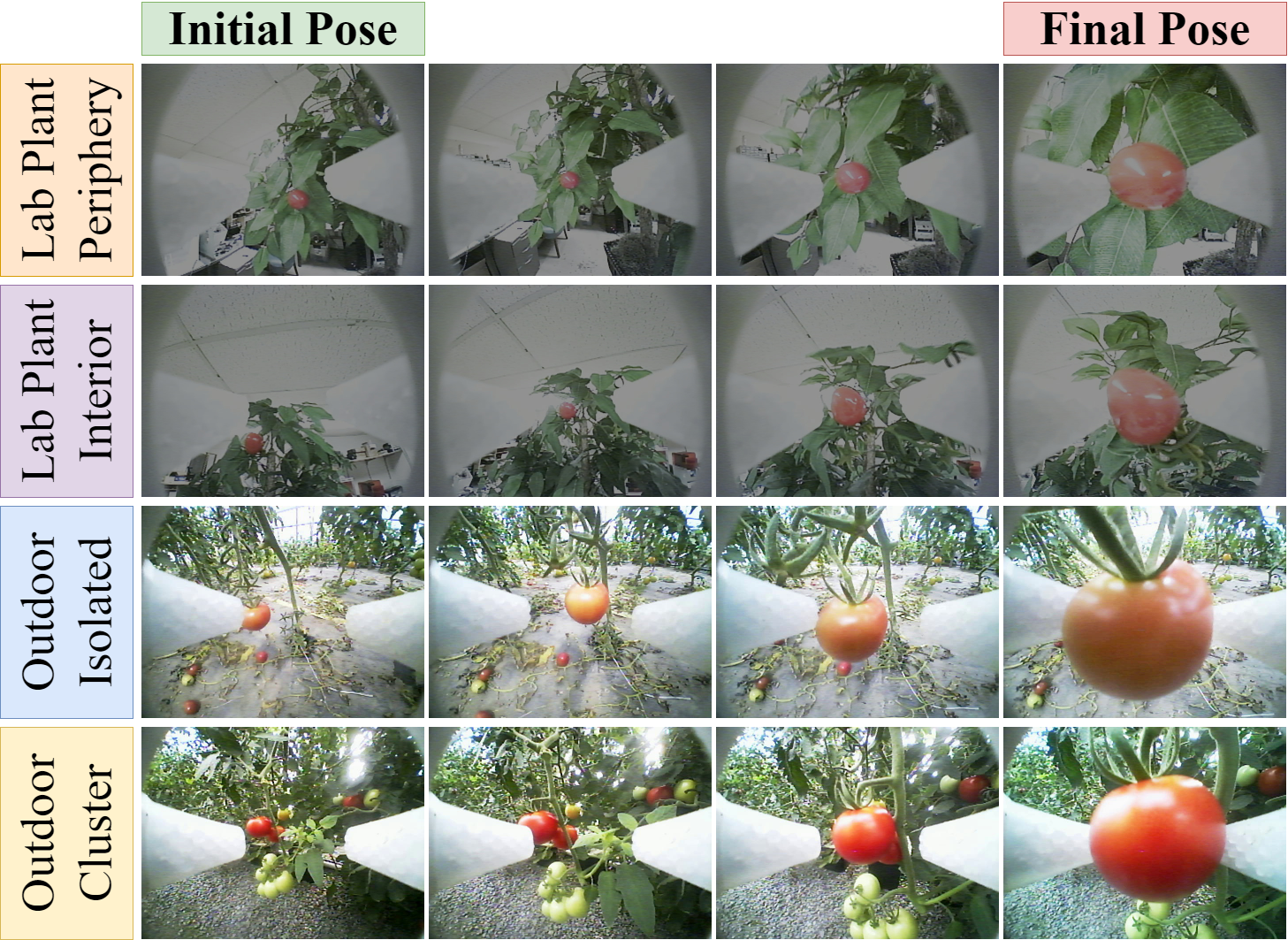}
  \caption{The images during visual servoing from the tip camera are presented for berries on the periphery of the plant as well as in the canopy. Outdoor images showing isolated and clustered fruit are also presented.}
  \label{fig:successful_experiments}
  \vspace{-.3cm}
\end{figure}
\section{DISCUSSION AND LIMITATIONS}

Experiments show that our system reliably and consistently reaches fruit in both peripheral and cluttered under-canopy settings. Furthermore, the visual servoing is robust to lighting and environment changes, while gracefully degrading in performance under heavily noised input. Specifically, in constrained and cluttered under-canopy settings, our system is shown to be more effective than larger manipulation systems with distal depth cameras. 

\begin{figure}[htp!]
\centering

\includegraphics[width=0.48\textwidth]{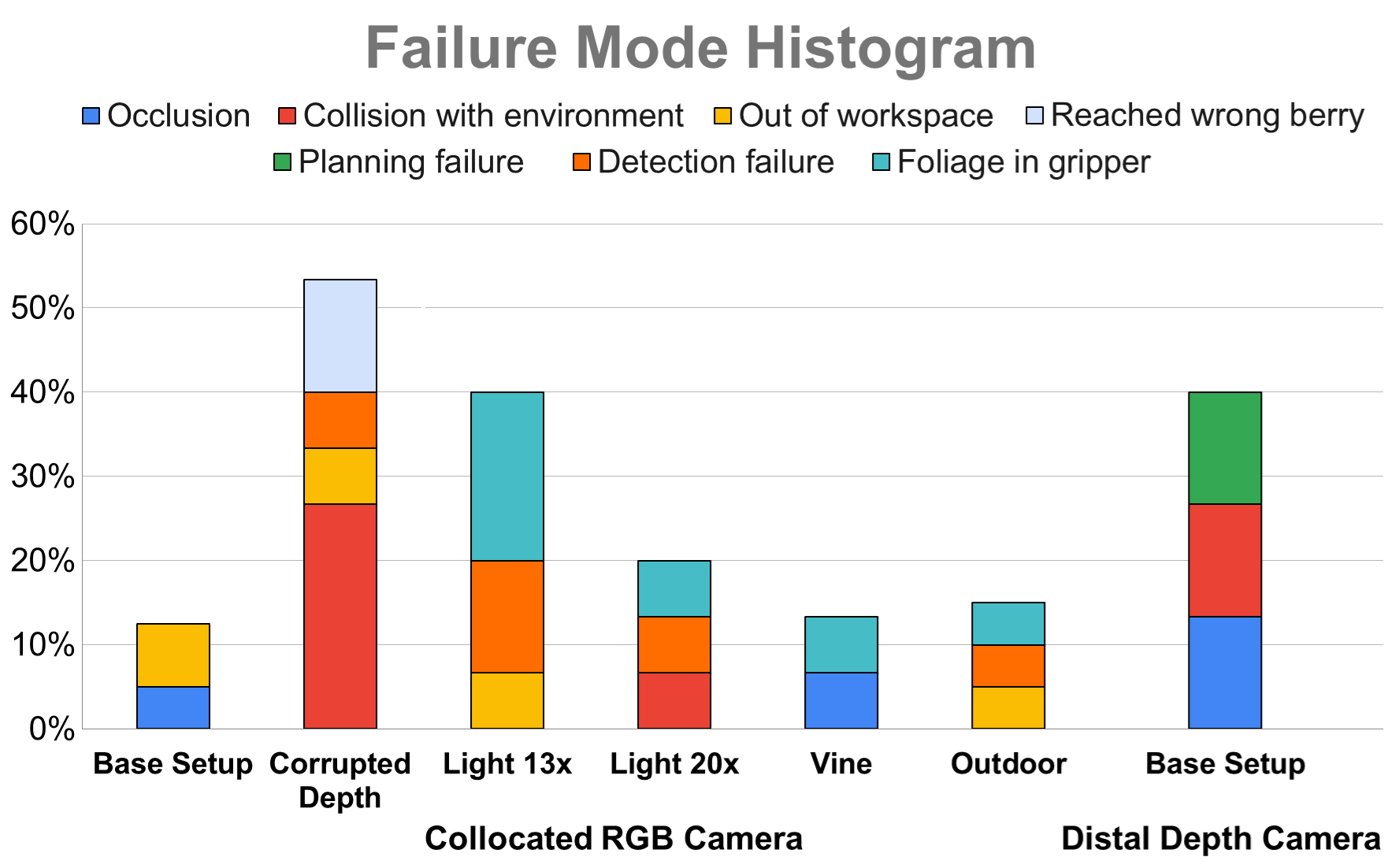}
  \vspace{-.6cm}
  \caption{The failure cases that occurred throughout testing. For each failure case, the frequency is broken down for each testing environment.}
  \label{fig:Failure_Cases}
  \vspace{-20pt}
\end{figure}

The success of our system depends on the accuracy of two serially connected subsystems, the global target detection, and the visual servoing scheme. Several different failure modes arise throughout the entire pipeline of our system. These are captured in Figure \ref{fig:Failure_Cases}. A prominent failure mode was \textit{occlusion of the target object} (marked as dark blue in Fig. \ref{fig:Failure_Cases}) during the approach phase of the manipulator, which occurred as the system unknowingly moved leaves or stems. This failure mode was exacerbated with the distal depth camera, where the gripper itself would often occlude the target. However, this failure mode did not occur in the outdoor setting as excess outer foliage was removed in preparation for harvest. 

The most frequent failure was \textit{collision of the manipulator link with the environment}. This occurred when there was a need to re-plan initial approach poses because the target was not in view mostly due to corrupted depth measurements from the base camera (seen as red in Figure \ref{fig:Failure_Cases}). In our setup, re-planning is performed close to the plant canopy and without the knowledge of plant occupancy in space thus increasing the chances of collision. As mentioned previously, the distal depth camera collided more with the branches and leaves leading to increased failures of this kind (see Figure \ref{fig:distal_failure}). This failure was not observed in the outdoor setting. 

Another important failure mode is \textit{the limited manipulator workpsace}, which occurs when the manipulator reaches the boundary of its workspace while centering the target with fixed orientation (shown as yellow in Figure \ref{fig:Failure_Cases}). This is due to the limitation of a compact 6-DOF manipulator and warrants anticipating collisions and re-planning the approach pose. Interestingly, this failure occurred most frequently in our baseline experiments and in the high tunnels indicating a mismatch between the field of view of the base camera and the safe workspace of the manipulator. 

\textit{Planning failures} uniquely occurred with the distal depth camera on the end effector. When reaching under the canopy and incorporating safety margins to prevent self-collision, our heuristics for computing initial poses could not find a safe pose.

\textit{Detection failures of fruit} occurred with different lighting scenarios. Interestingly, successful detection from the base camera did not guarantee successful detection from the tip camera despite being in view due to the location of the light source in indoor or outdoor settings. Despite their small profile, leaves and branches, along with the berry, can sometimes end up between the gripper fingers. This was considered a failure as would inhibited the downstream harvesting process. Harvesting ripe fruit requires minimal force, so grasps without obstruction would reliable lead to a successful pluck. Lastly, due to heavily noised depth measurements, the manipulator would approach a nearby fruit rather than the intended target.

\section{CONCLUSIONS AND FUTURE WORK}
In the quest for an automated robotic harvesting solution in space-constrained urban environments such as high tunnels and green houses, several hardware and software trade-offs are needed. For example, slender manipulator extremities preclude the use of large high-resolution depth cameras collocated with the end-effector reducing image richness that enables a fast grasp. Detect2Grasp offsets the hardware limitations by a visual servoing scheme that uses a combination of an RGB-D camera (global) stationed at the manipulator base and low-cost wireless RGB camera (local) at the manipulator tip. The depth camera scans the environment for ripe berries detected using the Yolov7 network. This information guides the manipulator and the end-effector camera to an approach pose with the target in view. Leveraging the visual feedback provided by the end effector camera, the manipulator iteratively centers the berry in the image space and approaches it to place it between the gripper. Our system was extensively evaluated on different test scenarios in a laboratory environment using artificial plants as well as outdoors using cherry tomatoes. High tunnel experiments showed an average reaching time of 10.98s with a 85.0\% success rate. 

As a result of this work, several future directions have been identified. First, constructing an occupancy grid of the environment would reduce collisions and occlusion problems. Second, modeling interactions with the plant could be incorporated by having additional sensors to avoid errors to a dynamic environment \cite{yao2023estimating}. Grasp quality could be enhanced with sensors or vision models to determine if foliage was between the grippers. Additionally, simultaneously scanning with the base and tip cameras could be integrated to detect berries faster from multiple perspectives. This would also improve recall of berries occluded from the base camera's perspective. 

\section*{ACKNOWLEDGMENT}
We would like to acknowledge iSEE Campus as a Living Lab seed funding, the Center for Digital Agriculture (CDA) at the University of Illinois at Urbana-Champaign, Dr. Kacie Athey for insightful discussions about the need for automation in high tunnels, as well as Dr. Justin Yim for insightful discussions about our results. We would also like to thank Nitin Nagarkar, Sameer Iyengar, and Everrett Wang for their work in helping develop the robotic system used in this work.

\bibliographystyle{IEEEtran}{}
\bibliography{references}
\end{document}